\pdfoutput=1
\documentclass[letterpaper, 10pt, conference]{ieeeconf/ieeeconf}    

\makeatletter
\let\NAT@parse\undefined
\makeatother

\usepackage[numbers,sectionbib,sort&compress]{natbib}

\usepackage{bm}
\usepackage{color}
\usepackage{gensymb}
\usepackage{graphicx}
\usepackage{amsmath}
\usepackage{algorithm}
\usepackage[noend]{algpseudocode}
\usepackage{subcaption}
\usepackage[font=small]{caption}
\usepackage[export]{adjustbox}
\usepackage{siunitx}
\usepackage{booktabs}
\usepackage{amsfonts}
\usepackage{natbib}
\usepackage[rightcaption]{sidecap} \sidecaptionvpos{figure}{c}

\usepackage{algorithmicx}
\algdef{SE}[DOWHILE]{Do}{doWhile}{\algorithmicdo}[1]{\algorithmicwhile\ #1}%

\usepackage{hyperref}
\hypersetup{colorlinks,breaklinks,
linkcolor=[rgb]{0.5,0.,0.},
citecolor=[rgb]{0.000,0.427,0.173},
urlcolor=[rgb]{0.031,0.318,0.612}}

\usepackage[nameinlink]{cleveref}
\usepackage{color}

\usepackage[printonlyused,withpage,nolist,nohyperlinks]{acronym}
\begin{acronym}

\acro{2D}{two-dimensional}

\acro{3D}{three-dimensional}

\acro{AHRS}{attitude and heading reference system}

\acro{AUV}{autonomous underwater vehicle}

\acro{CPP}{Chinese Postman Problem}

\acro{DoF}{degree of freedom}
\acrodefplural{DoF}[DoFs]{degrees of freedom}

\acro{DVL}{Doppler velocity log}

\acro{FSM}{finite state machine}

\acro{IMU}{inertial measurement unit}

\acro{LBL}{Long Baseline}

\acro{MCM}{mine countermeasures}

\acro{MDP}{Markov decision process}
\acrodefplural{MDP}[MDPs]{Markov decision processes}

\acro{POMDP}{partially observable Markov decision process}
\acrodefplural{POMDP}[POMDPs]{partially observable Markov decision processes}

\acro{PRM}{Probabilistic Roadmap}
\acrodefplural{PRM}[PRM]{Probabilistic Roadmaps}

\acro{ROI}{region of interest}
\acrodefplural{ROI}[ROIs]{regions of interest}

\acro{ROS}{Robot Operating System}

\acro{ROV}{remotely operated vehicle}

\acro{RRT}{Rapidly-exploring Random Tree}
\acrodefplural{RRT}[RRTs]{Rapidly-exploring Random Trees}

\acro{SLAM}{simultaneous localization and mapping}

\acro{SSE}{sum of squared errors}

\acro{STOMP}{Stochastic Trajectory Optimization for Motion Planning}

\acro{TRN}{Terrain-Relative Navigation}

\acro{UAV}{unmanned aerial vehicle}

\acro{USBL}{Ultra-Short Baseline}

\acro{IPP}{informative path planning}

\acro{FoV}{field of view}
\acrodefplural{FoV}[FoVs]{fields of view}

\acro{CDF}{cumulative distribution function}

\acro{ML}{maximum likelihood}

\end{acronym}

\definecolor{CommentPink}{rgb}{1,0.2,0.5}

\usepackage[normalem]{ulem}

\IEEEoverridecommandlockouts                           
\title{\LARGE \bf
Obstacle-aware Adaptive Informative Path Planning \\ for UAV-based Target Search
}

\author{$\text{Ajith Anil Meera\textsuperscript{1}}$, $\text{Marija Popovi\'{c} \textsuperscript{2}} $, $\text{Alexander Millane \textsuperscript{3}}$ and $\text{Roland Siegwart\textsuperscript{4}}$ \thanks{\textsuperscript{1,2,3,4} are with
the Autonomous Systems Lab. (ASL), ETH Z\"{u}rich, Z\"{u}rich, 
Switzerland. Author \textsuperscript{1} is also affiliated with TU Delft, The Netherlands. Corresponding author: \texttt{mpopovic@ethz.ch}.}
}

\begin{document}

\maketitle
\thispagestyle{empty}
\pagestyle{empty}

\begin{abstract}
Target search with unmanned aerial vehicles (UAVs) is relevant problem to many scenarios, e.g., search and rescue (SaR).
However, a key challenge is planning paths for maximal search efficiency given flight time constraints.
To address this, we propose the Obstacle-aware Adaptive Informative Path Planning (OA-IPP) algorithm for target search in cluttered environments using UAVs.
Our approach leverages a layered planning strategy
using a Gaussian Process (GP)-based model of target occupancy
to generate informative paths in continuous 3D space.
Within this framework, 
we introduce an adaptive replanning scheme
which allows us to trade off between information gain, field coverage, sensor performance, and collision avoidance for efficient target detection.
Extensive simulations show that our OA-IPP method performs better than state-of-the-art planners,
and we demonstrate its application in a realistic urban SaR scenario.
\end{abstract}

\section{INTRODUCTION} \label{S:introduction}
Autonomous target search in cluttered environments is a challenging problem relevant for a wide range of applications, e.g., finding victims in SaR operations \cite{waharte2010supporting,gupta2017decision}, monitoring vegetation in precision agriculture \cite{popovic2017multiresolution, Colomina2014}, patrolling military borders \cite{Girard2004}, and tracking endangered species \cite{Linchant2015}. With recent technological advances, UAVs are rapidly gaining popularity as a aerial data acquisition tool for this task. Compared with traditional approaches, such as manned aircraft and ground-based search, they offer high maneuverability, adaptability, and can provide data with high spatial and temporal resolution at a lower cost \cite{Colomina2014}.


However, deploying a UAV to search for targets presents several challenges.
A fundamental task is to plan the robot's motion to maximize the information collected about a terrain given its resource constraints, such as finite battery life.
For detection with an on-board camera, a key consideration is that area coverage can be achieved at different resolutions depending on the flying altitude.
Finally, in cluttered environments, e.g., urban scenarios, it must be ensured that the generated paths are obstacle-free with minimal obstructions to the camera field of view (FoV).
%
\begin{figure}[htpb]
    \centering
    \captionsetup{justification=justified}
    \begin{subfigure}[b]{0.25\textwidth}
		\includegraphics[width=\textwidth]{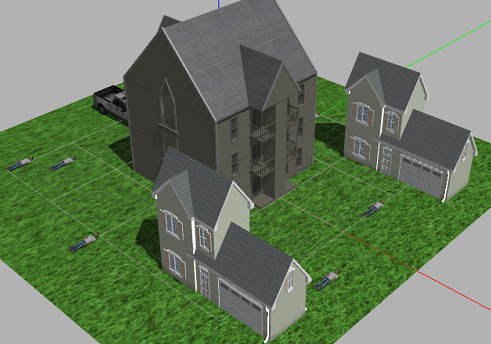}
		\caption{Isometric view.}
		\label{fig:envt_real_iso}
    \end{subfigure} 
    \begin{subfigure}[b]{0.13\textwidth}
		\includegraphics[width=\textwidth]{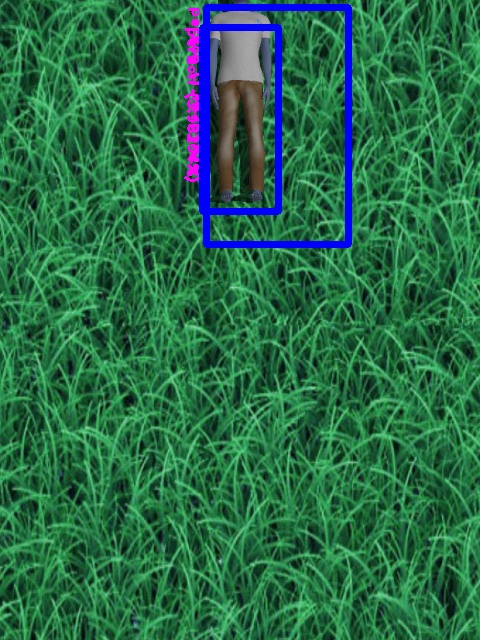}
		\caption{Detection.}
		\label{fig:GP_YOLO_5m}
    \end{subfigure} 
    \begin{subfigure}[b]{0.22\textwidth}
		\includegraphics[width=\textwidth]{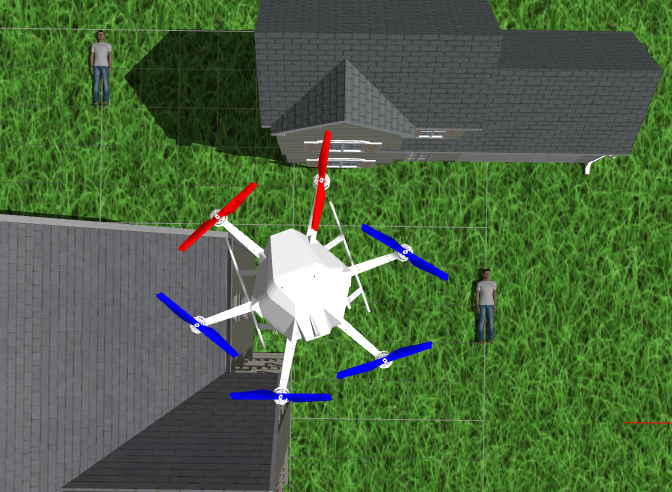}
		\caption{Scouting UAV.}
		\label{fig:top_view_UAV}
    \end{subfigure} 
    \begin{subfigure}[b]{0.16\textwidth}
		\includegraphics[width=\textwidth]{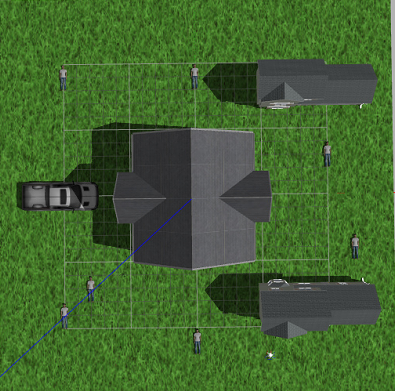}
		\caption{Top view.}
		\label{fig:envt_real_top}
    \end{subfigure}   
     \begin{subfigure}[b]{0.2\textwidth}
		\includegraphics[width=\textwidth,trim={.45cm .45cm 0 0},clip]{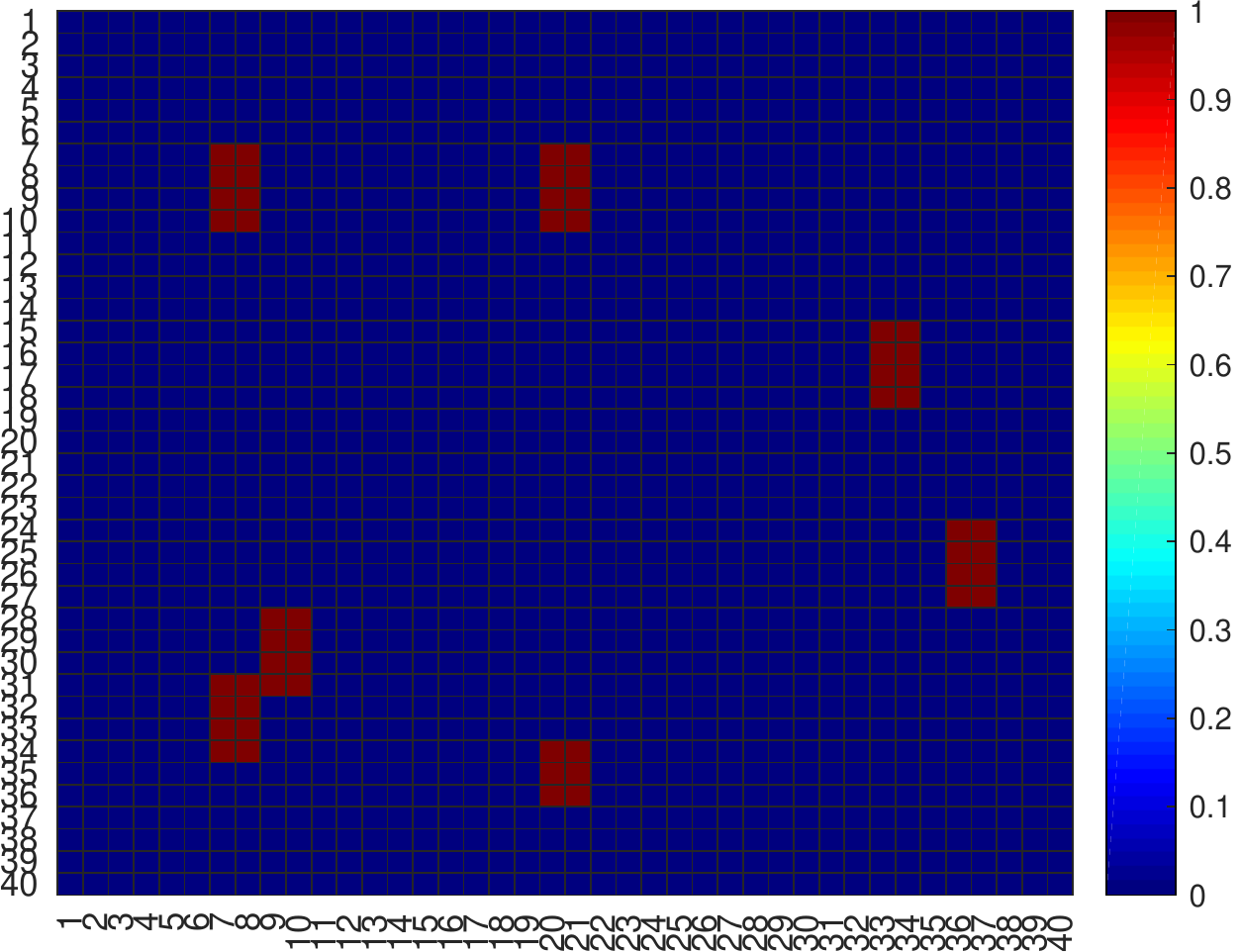}
		\caption{Ground truth.}
		\label{fig:realistic_ground_truth}
    \end{subfigure} 
    \begin{subfigure}[b]{0.2\textwidth}
		\includegraphics[width=\textwidth,trim={.45cm .45cm 0 0},clip]{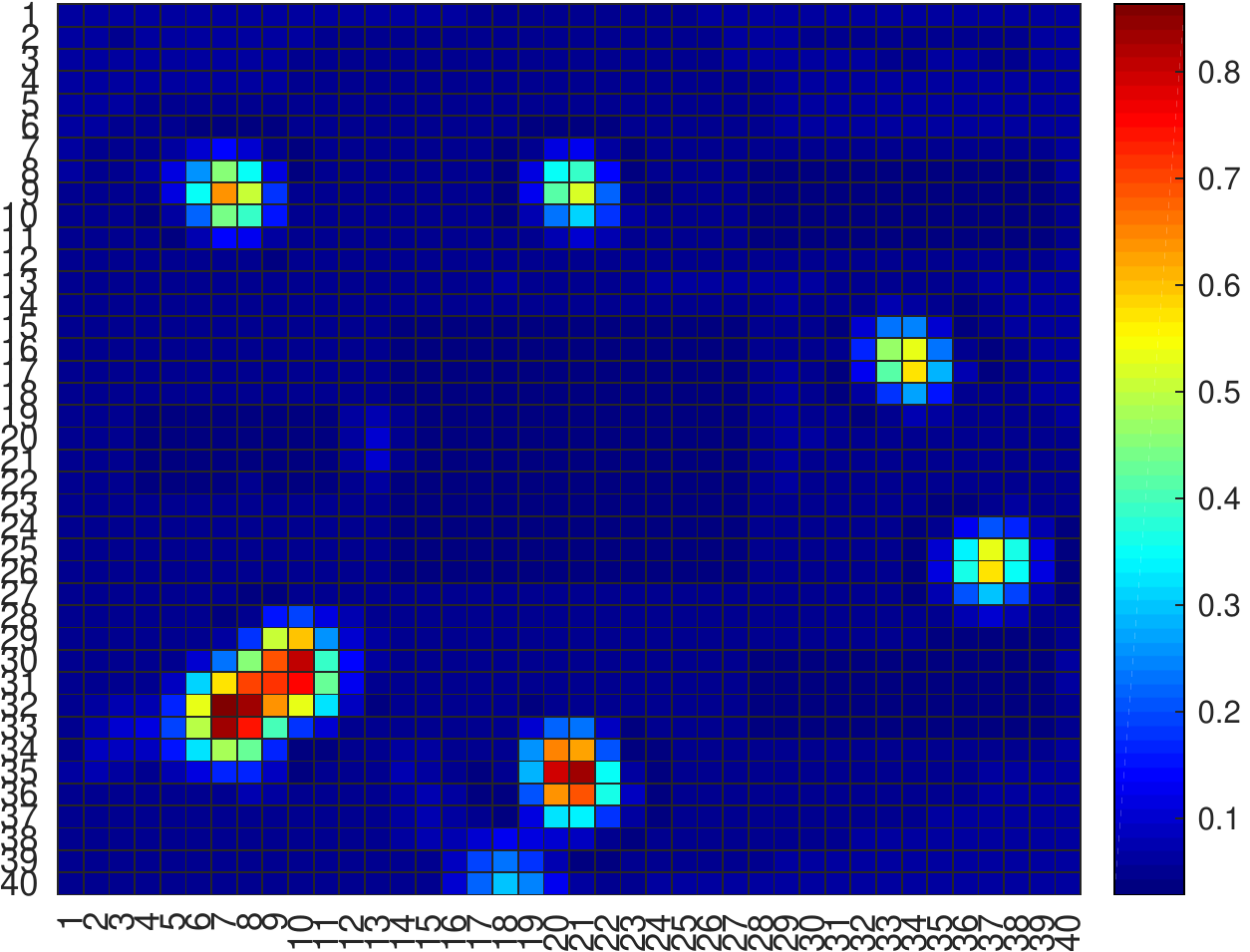}
		\caption{Final map after 150s.}
		\label{fig:realistic_final_mean}
    \end{subfigure} 
    \caption{(a,c,d) depicts our realistic Gazebo-based simulation of an urban search and rescue (SaR) scenario with an unmanned aerial vehicle (UAV) scouting for victims on the field.
    (b) exemplifies a target victim detection using the on-board camera. (e) depicts the ground truth for human occupancy and (f) depicts the target map after the UAV flight. All targets are correctly detected.}
    \label{fig:realistic_results}
\end{figure}


%
In this work we propose an algorithm that tackles these issues simultaneously.
Our problem formulation considers a cluttered, known environment in which the UAV navigates.
In this set-up,
the objective is to quickly find targets on a ground field using an on-board camera. 
Our framework consists of three major components:
(1) modelling of the 3D environment;
(2) mapping of the target 2D terrain using a probabilistic sensor model;
and (3) planning.
The main idea is to treat target search as an optimization process
that solves an IPP problem in continuous 3D space
and couples the aspects presented above.
This allows us to generate smooth, obstacle-free paths
for efficient data acquisition, which abide by a limited time budget.
Moreover, within our approach,
we introduce an adaptive planning strategy
to quickly focus on areas where targets are likely to be found.

The core contributions of this work are:
\begin{enumerate}
\item an algorithm for UAV-based target search which couples informative planning with obstacle awareness,
\item an adaptive replanning scheme based on Bayesian Optimization (BO) which trades off between exploration and exploitation for efficient target detection, and
\item the extensive evaluation of our framework in simulation and its validation in a realistic urban SaR scenario.
\end{enumerate}

Figure \ref{fig:realistic_results} demonstrates the success of our algorithm in detecting all human targets on the field for an SaR problem. 



\section{RELATED WORK} \label{S:related_work}
The task of searching for targets is relevant to many real-world scenarios \cite{gupta2017decision,geyer2008active,Colomina2014,Linchant2015}.
Most generally,
this problem can be expressed as a Partially Observable Markov Decision Process (POMDP),
which models planning under uncertainty.
However, despite recent advances,
current POMDP solvers \cite{waharte2010supporting,Kurniawati2008}
still scale poorly for practical applications due to high-dimensional belief space.

In robotics, target search is often more efficiently formulated as an IPP problem,
where information about targets is maximized subject to a budget constraint.
IPP literature can broadly be classified based on three aspects:
(i) adaptivity; (ii) myopicity; and (iii) continuity.
Adaptive IPP \cite{popovic2017online} differs from non-adaptive IPP \cite{binney2013optimizing} in terms of how new measurements influence the planning routine.
The former performs online replanning
while the latter executes an \textit{a \ priori} path.
Myopic methods \cite{singh2010modeling} differ from non-myopic ones \cite{popovic2017multiresolution} in terms of their planning look-ahead.
The former navigates greedily by selecting next-best views,
while the latter plans with a finite lookahead to escape local minima.
Finally, whereas discrete strategies \cite{binney2013optimizing,binney2012branch}
perform combinatorial optimization on pre-defined grids,
continuous solvers \cite{popovic2017multiresolution,hollinger2014sampling,marchant2014bayesian} operate directly in the robot workspace to achieve better scalability.

Our algorithm is adaptive, non-myopic, and continuous.
It falls into a recently emerged category of methods which optimize a smooth continuous trajectory for maximal information gain \cite{popovic2017multiresolution,popovic2017online,brochu2010tutorial,marchant2014bayesian}.
Specifically,
we build upon the work of \citet{popovic2017multiresolution}
by introducing a new adaptive planning scheme based on BO \cite{brochu2010tutorial}
to perform exploration while focusing on potential targets as they are detected.


Accounting for sensor uncertainty and facilitating target re-observation is crucial in procuring an error-free target map. Despite abundant prior research, accounting for both of these factors in cluttered environments is an understudied topic.
Although \citet{dang2018autonomous} and \citet{marchant2014bayesian} plan for target re-observation, they do not consider variable sensor performance. Whereas the methods of \citet{popovic2017multiresolution} cater for sensor uncertainties, they do not address obstacles.
By unifying these concepts,
our work aims to bridge the gap towards practical applications, i.e., urban~\cite{geyer2008active,richter2016polynomial} and natural~\cite{waharte2010supporting} settings.

\section{PROBLEM STATEMENT} \label{S:problem_statement}
We formulate the target search task as an IPP problem as follows. Our aim is to find an optimal continuous trajectory $\psi$ in the space of all trajectories $\Psi$ for the robot:
\begin{equation} \label{E:IPP_obj}
\psi^* = \operatorname*{arg\,max}_{\psi \in \Psi} \ \frac{k_1 O_{info}(\psi)-k_2 C_{coll}(\psi)}{t_{flight}(\psi)}\textit{,}
\end{equation}
where $O_{info}(\psi)$, $C_{coll}(\psi)$, and $t_{flight}(\psi)$ are functions quantifying the information quality, collision cost, and flight time along a trajectory $\psi$, and $k_1$ and $k_2$ are non-negative constants trading off between useful data acquisition and obstacle avoidance.

\section{PRELIMINARIES} \label{S:modeling}

To lay the foundation for our IPP framework,
this section describes our methods of modeling the key elements of the target search problem.
We first detail our method of representing cluttered flying environments
before describing our field mapping strategy.


\subsection{Environment modeling} \label{S:envt_modeling}
Our set-up considers a known cluttered 3D environment above the monitored terrain,
in which the UAV flies.
We treat obstacles as standard geometric shapes and use the Voxblox system~\citep{oleynikova2017voxblox}, based on the Euclidean Signed Distance Function (ESDF), to enable computationally inexpensive collision checks. A hard penalty for colliding with obstacles is placed
based on an artificial potential field, defined as:
\begin{equation}
\label{E:hard_APF}
\ C_{hard}(\textbf{x})=
\begin{cases}
    0 & \text{if } \text{ESDF}(\textbf{x})\geq \frac{r_{UAV}}{2} \textit{,}\\
    1              & \text{otherwise} \textit{,}
\end{cases}
\end{equation}
where ESDF(\textperiodcentered) is the signed distance to the closest obstacle from a UAV configuration $\textbf{x}$, and $r_{UAV}$ is the radius of the smallest sphere that can contain the UAV.

Based on Equation~\ref{E:hard_APF}, we define the collision cost $C_{coll}(\psi)$ in Equation~\ref{E:IPP_obj} as:
\begin{equation} \label{E:coll_cost}
 C_{coll}(\psi) = \sum_{\textbf{x} \in \psi} C_{hard}(\textbf{x}) \textit{.} 
\end{equation}
Note that $C_{coll}(\psi)$ is non-zero if any point $\textbf{x} \in \psi$ is inside the obstacle, thereby incurring a penalty to the objective function. $C_{coll}(\psi)$ is evaluated by sub-sampling $\psi$ at a high sampling frequency for flight safety.

\subsection{Field modeling and mapping} \label{S:field_modeling}
Our field mapping strategy is based on the approach of~\citet{popovic2017multiresolution},
allowing us to incorporate probabilistic sensor models for data fusion
with constant-time measurement updates.
We represent the monitored ground field of target occupancy using a GP model,
which captures spatial correlations in a probabilistic and non-parametric way \cite{Rasmussen:2005:GPM:1162254}.
The field is assumed to be a continuous function in 2D space $\zeta: \varepsilon \rightarrow \mathbb{R}$, where $\varepsilon \subset \mathbb{R}^2$ is a location on the ground field. The GP is fully characterized by the mean $\mu = E[\zeta]$ and the covariance $P=E[(\zeta - \mu)(\zeta^T-\mu^T)]$
as $\zeta \sim GP(\mu,P)$, where $E[$\textperiodcentered$]$ is the expectation operator.

The field is discretized at a given resolution to obtain $n$ training locations $X\in \epsilon$. A set of $n^\prime$ prediction points $X^\prime\in \epsilon$ are specified at which prior map is to be inferred.
To describe the field,
we propose using the isotropic Mat\'ern 3/2 kernel function \cite{Rasmussen:2005:GPM:1162254} common in geostatistical analysis, due to its capability to capture discrete targets.
It is defined as:
\begin{equation} \label{E:matern}
k_{Mat3}(x,x^\prime) = \sigma_f^2\bigg(1+\frac{\sqrt{3}d}{l}\bigg) \exp\bigg(-\frac{\sqrt{3}d}{l}\bigg) \textit{,}
\end{equation}
where $l$ and $\sigma_f^2$ are the hyperparameters representing the lengthscale and signal variance, respectively, and $d$ is the Euclidean distance between input locations $x$ and $x^\prime$.

The covariance is evaluated using:
\begin{align}
\label{eqn_covariance}
P ={}& K(X^\prime,X^\prime)-K(X^\prime,X)[K(X,X)+\sigma_n^2 I]^{-1} \times &\notag\\
&K(X^\prime,X)^T \textit{,}
\end{align}
where $K(X,X^\prime)$ denotes the $n \times n^\prime$ matrix of the covariances evaluated at all pairs of training and test points, $P$ is the posterior covariance, and $\sigma_n^2$ is a hyperparameter representing the noise variance. These hyperparameters $\{l, \sigma_f^2,\sigma_n^2\}$ can be learned by maximizing the log marginal likelihood of a training dataset~\cite{Rasmussen:2005:GPM:1162254}.

Sensor measurements are fused sequentially into the GP field map in two steps:
(1) FoV estimation and (2) data fusion.
First,
the camera FoV at the measurement pose is projected to index the cells in $X$ to be updated.
In this step, we account for occlusions arising from obstacles in the 3D ESDF environment (Section~\ref{S:envt_modeling}).
Then, the field map is updated using a recursive technique based on the Kalman Filter (KF),
presented by~\citet{popovic2017multiresolution}.
Within our framework,
this method enables capturing the altitude-dependant performance of practical target detectors for probabilistic mapping,
as discussed in the following section.

\subsection{Sensor modeling} \label{S:sensor_modeling}
In this section,
we develop a sensor model for planning
to exemplify the integration of our framework with a real target detector.
Specifically,
we consider You Only Look Once (YOLO) \cite{redmon2016you},
a typical neural network-based detector using images,
for the application of SaR.

\begin{SCfigure}[][!h]
\centering
\includegraphics[width = 0.62\columnwidth]{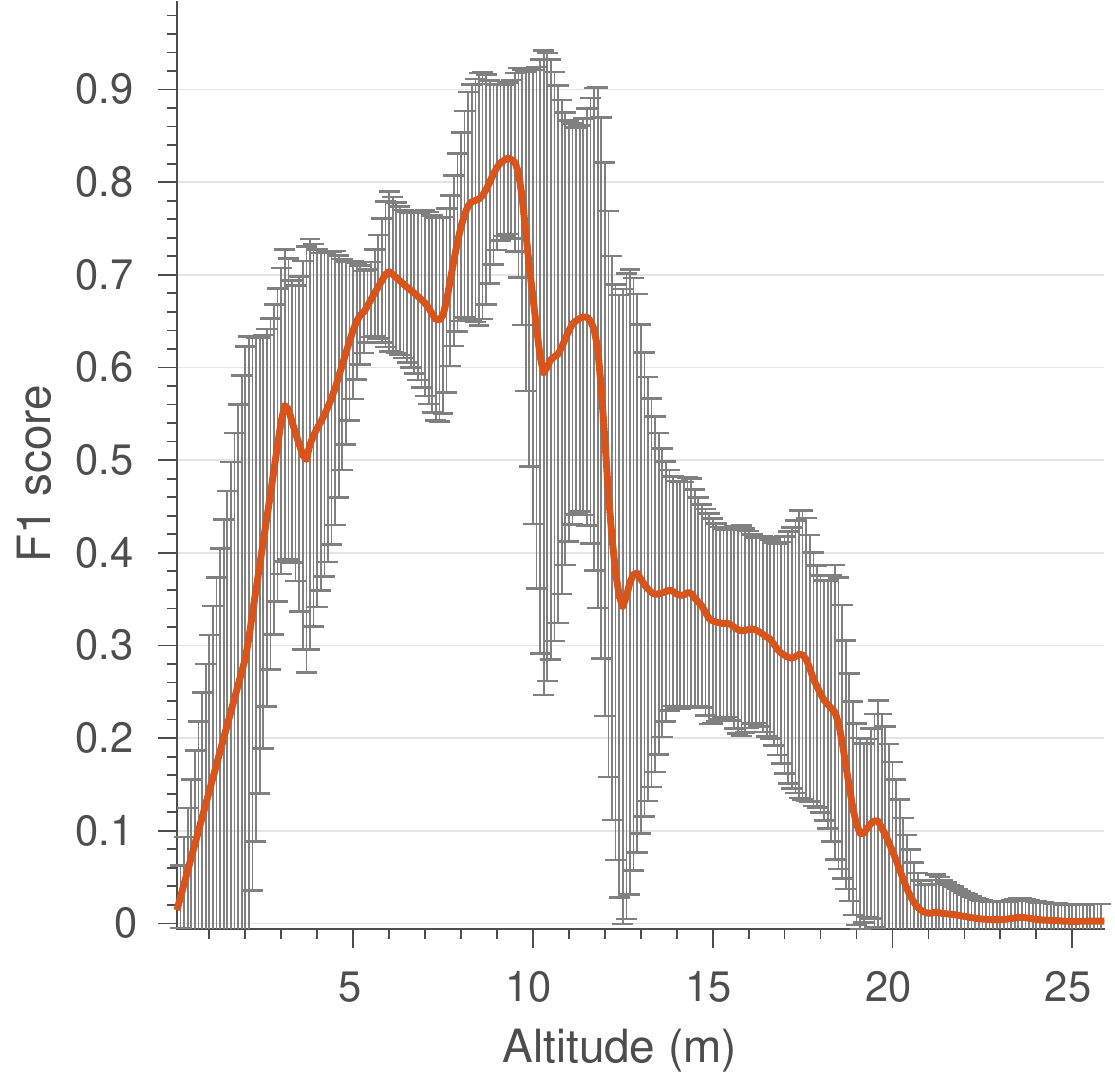}
\caption{Empirical analysis of accuracy for the YOLO human detector over a range of flying altitudes.
The orange curve depicts the mean over 100 detections,
and the gray error bars correspond to one standard deviation. A variable sensor performance with respect to altitude can be observed.}
\label{fig:F1_score}
\end{SCfigure}

To investigate how detection accuracy varies with FoV,
the performance of YOLO is evaluated empirically using multiple images simulated at different altitudes,
each containing one human target.
We quantify accuracy using F1 score by counting successful target detections.
Figure \ref{fig:F1_score} summarizes our results for an altitude range of $0-25$\,m.
This plot suggests that YOLO performs best at an intermediate altitude of $\sim 10$\,m,
with an altitude of $\sim 20$\,m beyond which accuracy deteriorates significantly due to coarse image resolution, larger FoV and higher clutter in image.
This analysis motivates our planner
in Section~\ref{S:adaptive_planning}
which caters for altitude dependency.

\subsection{Global Optimization}
We use the state-of-the-art evolutionary optimizer called CMA-ES \cite{hansen2006cma} for global optimization of Equation \ref{E:IPP_obj}. The successful application of CMA-ES in gradient-free, high-dimensional and non-linear optimization problems \cite{popovic2017multiresolution,hitz2017adaptive}, along with its quasi-parameter free nature, motivate its selection when compared to other global optimizers \cite{hansen2006cma}.


\section{PLANNING} \label{S:planning}
In this section,
we present
our planning framework,
named the Obstacle-aware Adaptive Informative Path Planner (OA-IPP) algorithm.
By coupling information gathering with obstacle avoidance,
our method achieves efficient target search in cluttered environments. The planner is an extension of \citet{popovic2017multiresolution}, enabling it with obstacle-aware features and introducing a layered optimization for exploration-exploitation trade-off. 

\subsection{Trajectory parametrization}
A polynomial trajectory $\psi$ is parameterized by a sequence of $N$ control waypoints to be visited by the UAV, defined as $C=[c_1,...,c_N]$, where the first waypoint $c_1$ represents the current UAV location. The polynomial trajectory connects these control points using $N-1$ $k$-order spline segments for minimum-snap dynamics as given by~\citet{richter2016polynomial}. Along $\psi$,
we consider a constant frequency for the sensor, computing the spacing between measurement poses with respect to the UAV dynamics.

\subsection{Algorithm} \label{S:algorithm}
Algorithm \ref{algo:init_control_pose} calculates the initial greedy solution $C_{init}$ comprising of waypoints $nbvp$, based on the current UAV pose $R_0$. $nbvp$ is the next best viewpoint for the UAV which is estimated by randomly sampling points from the environment at different altitudes and greedily maximizing the IPP objective given in Equation \ref{E:IPP_obj} (Line \ref{A:line:nbvp}). Additionally, each waypoint should be visible from the previous viewpoint for a collision free initial solution. The next waypoint is evaluated by assuming a measurement update at the previous waypoint (Line \ref{A:line:update_nbvp_map}).

\begin{algorithm}
\caption{Function to perform the greedy search.}\label{algo:init_control_pose}
\begin{algorithmic}[1]
\Function{\textsc{CoarseGreedySearch}($R_0$)}{}
\State Initialize empty queue $C_{init}$.
\State Insert $R_0$ into $C_{init}$.
\State Initialize $GP\_sim$ to $GP$.
\For {$i=1:N-1$}
\State $nbvp = \textsc{NextBestViewPoint}(C_{init}[i],GP\_sim)$\label{A:line:nbvp}
\State Insert $nbvp$ into $C_{init}$.
\State Update $GP\_sim$ map at $nbvp$.\label{A:line:update_nbvp_map}
\EndFor
\State \Return $C_{init}$
\EndFunction
\end{algorithmic}
\end{algorithm}

Algorithm \ref{algo:IPP_SaR} summarizes our approach. The UAV evaluates and initial path $C_{init}$ (Line \ref{A:line:greedy}) and refines it using global optimization (CMAES) to generate path $C$ (Line \ref{A:line:globopt}), which is checked for collisions (Line \ref{A:line:coll_check}). The path $C$ is used for data acquisition and mapping (Line \ref{A:line:flight_start}-\ref{A:line:flight_end}) at a constant measurement frequency. A new path is planned once the stipulated path flight time is complete. This is repeated until the budget flight time is exhausted.

\begin{algorithm}
\caption{\textsc{OA-IPP} routine.} \label{algo:IPP_SaR}
\begin{algorithmic}[1]
\State Create ESDF map.
\State Initialize the GP.
\If {$t_{flight}<Budget$} \Comment{Replanning.}
\Do
\State $C_{init} = \textsc{CoarseGreedySearch}(R_0)$. \label{A:line:greedy}
\State $C=\textsc{GlobalOptimization}(C_{init})$  \label{A:line:globopt}
\doWhile{not $\textsc{isCollisionFreePath}(C,ESDF)$}\label{A:line:coll_check}
\Do \label{A:line:flight_start}
\State Fly through $C$ and take measurements.
\State Update $GP$ map at measurement locations.
\doWhile{flight time for $C$ not exhausted} \label{A:line:flight_end}
\State Stop UAV and hover.
\Else \State {Land the UAV.}
\EndIf
\end{algorithmic}
\end{algorithm}

\subsection{Uncertainty reduction as the objective} \label{S:nonadaptive_planning}
Many IPP approaches use uncertainty reduction as the objective for the IPP problem without using the target detections for planning \cite{popovic2017online,popovic2017multiresolution}. The information gain can be represented as:
\begin{equation} \label{eqn:var_reduction}
O_{info}(\psi) = Tr(P^-)-Tr(P^+),
\end{equation}
where $Tr(.)$ denotes the trace of the matrix, $P^-$ and $P^+$ are the prior and posterior covariances, respectively. The latter is evaluated by fusing all measurements along $\psi$. However, the planner is explorative and does not take target detection into account while planning.

\subsection{Layered optimization for adaptive planning} \label{S:adaptive_planning}
In this section we propose a layered optimization approach that facilitates target re-observation, rendering the planner with exploration-exploitation trade-off capability, which is crucial for robustness against wrong detections. We define an information-theoretic objective function based on the acquisition function that is used in the BO framework. We use the Upper Confidence Bound (UCB) \cite{cox1992statistical} defined by Equation \ref{E:ucb} as the acquisition function.

\begin{equation} \label{E:ucb}
UCB(x) = \mu(x)+\kappa \sigma(x),
\end{equation}
Here $\kappa$ is the exploration-exploitation tuning parameter, and $\mu$ and $\sigma^2$ are the mean and variance of the GP. We define Acquisition View (AV) as the sum total of acquisition function values within the FoV, given by 
\begin{equation}
AV=\sum_{x\in FoV} UCB(x).
\end{equation}
The sensor performance curve given in Figure \ref{fig:F1_score} is modelled as a normal distribution $N(h_{opt},\sigma_1)$, and is used as a reward function for the objective function. The information gain is then defined as the combination of sensor performance and AV as
\begin{equation} \label{eqn:obj_acqu}
O_{info} =
\begin{cases}
 AV \times \Bigg[\frac{1}{\sigma_1 \sqrt{2\pi}}e^{-\frac{1}{2}\big(\frac{h-h_{opt}}{\sigma_1} \big)^2} \Bigg] & h<h_{sat},\\
0 & \text{otherwise}.
\end{cases}
\end{equation} 
The AV increases with altitude, while the sensor performance follows the pattern in Figure \ref{fig:F1_score}. Therefore, this layer of initial optimization based on BO given by Equation \ref{eqn:obj_acqu} encodes the trade-off between sensor performance and FoV by facilitating target re-observation through a balanced exploration-exploitation strategy that is embedded within the UCB. This objective contributes to the next optimization layer to evaluate the optimal 3D path.

\section{EXPERIMENTAL RESULTS} \label{S:experimental_results}
This section presents our experimental results. We validate our framework in simulation by comparing it to state-of-the-art methods and study the effects of our adaptive replanning scheme. We then show its application in a realistic urban SaR scenario.

\subsection{Benchmarking} \label{S:benchmarking}
We evaluate our algorithm on a $30 \times 30 \times 26$\,m target search scenario in the RotorS-based simulation environment \cite{Furrer2016}. The set-up, shown in Figure \ref{fig:envt_men_one_side}, features a $4 \times 10 \times 26$\,m obstacle and 7 human targets. Targets are placed on lower half of the field, necessitating target re-observation in that region. We use a uniform resolution of $0.75$\,m for the GP model, and initialize the mean with a constant low prior of 0.1, assuming a sparse target distribution. The hyperparameters of the Mat\'ern 3/2 kernel function are similar to the training results given in \cite{popovic2017multiresolution}: $\{\sigma_n^2,\sigma_f^2,l\}=\{1.42,1.82,3.67\}$.
\begin{figure}[htpb]
    \centering
    \begin{subfigure}[b]{0.2\textwidth}
		\includegraphics[width=\textwidth]{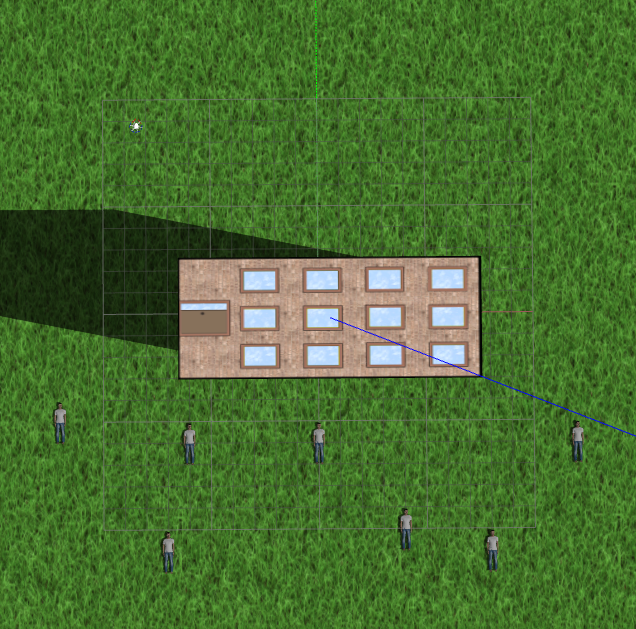}
		\caption{}\label{fig:envt_men_one_side}
    \end{subfigure} 
    \begin{subfigure}[b]{0.278\textwidth}
		\includegraphics[width=\textwidth,trim={1.8cm 1.8cm 0 0},clip]{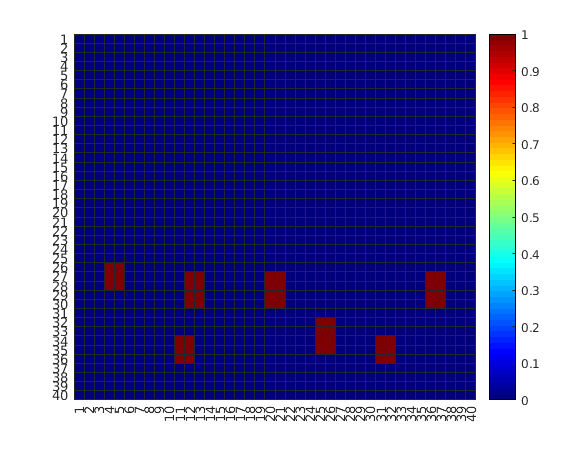}
		\caption{}\label{fig:rand_box_big_build}
    \end{subfigure}   
    \caption{(a) shows our simulation set-up in RotorS featuring a tall building and 7 human targets placed on lower half of the field. (b) visualizes the ground truth map of human occupancy. }\label{fig:envt_men_one_side}
\end{figure}

We use YOLO Tiny 2.0 for human detection with a threshold value of 0.05 and measurement frequency of $0.15$\,Hz. The detector receives images from a downward facing camera on the UAV with a FoV of $(45, 60)\degree$. For planning, the constants in Equation \ref{eqn:obj_acqu} were experimentally computed as: $h_{opt} = 10$\,m, $h_{sat} = 26$\,m and $\sigma_1 = 7$. The parameters of the altitude dependent uncertainty model given in \cite{popovic2017multiresolution} are $(A, B)=(1, 0.05)$.

We compare our approach against (1) "lawnmower" coverage planning \cite{galceran2013survey}
and (2) random waypoint selection,
running 25 trials for each method with a flight budget of $150$\,s.
To evaluate performance,
we quantify accuracy using the Root Square Error (RSE) with respect to the ground truth map of human occupancy
with the same resolution as the GP.
We opt for this metric as it weighs false positive and negative misclassifications equally,
reflecting the practical aims of a SaR mission.
During a mission,
higher rates of RSE reduction indicate more efficient search performance.

For the coverage planner, a fixed flight altitude of $10$\,m was set
by considering various ``lawnmower" patterns for complete coverage
within the time budget and selecting the best one.
For the random planner,
we uniformly sample a destination measurement pose above the field and generate a trajectory by connecting it to the current UAV position.
For our planner, we set a reference speed and acceleration of $5$\,m/s and $3$\,m/s$^2$ for trajectory optimization \cite{richter2016polynomial} with the adaptive planning strategy from Section~\ref{S:adaptive_planning}.

\begin{SCfigure}[][htpb]
\centering
\captionsetup{justification=justified}
\includegraphics[scale = 0.35]{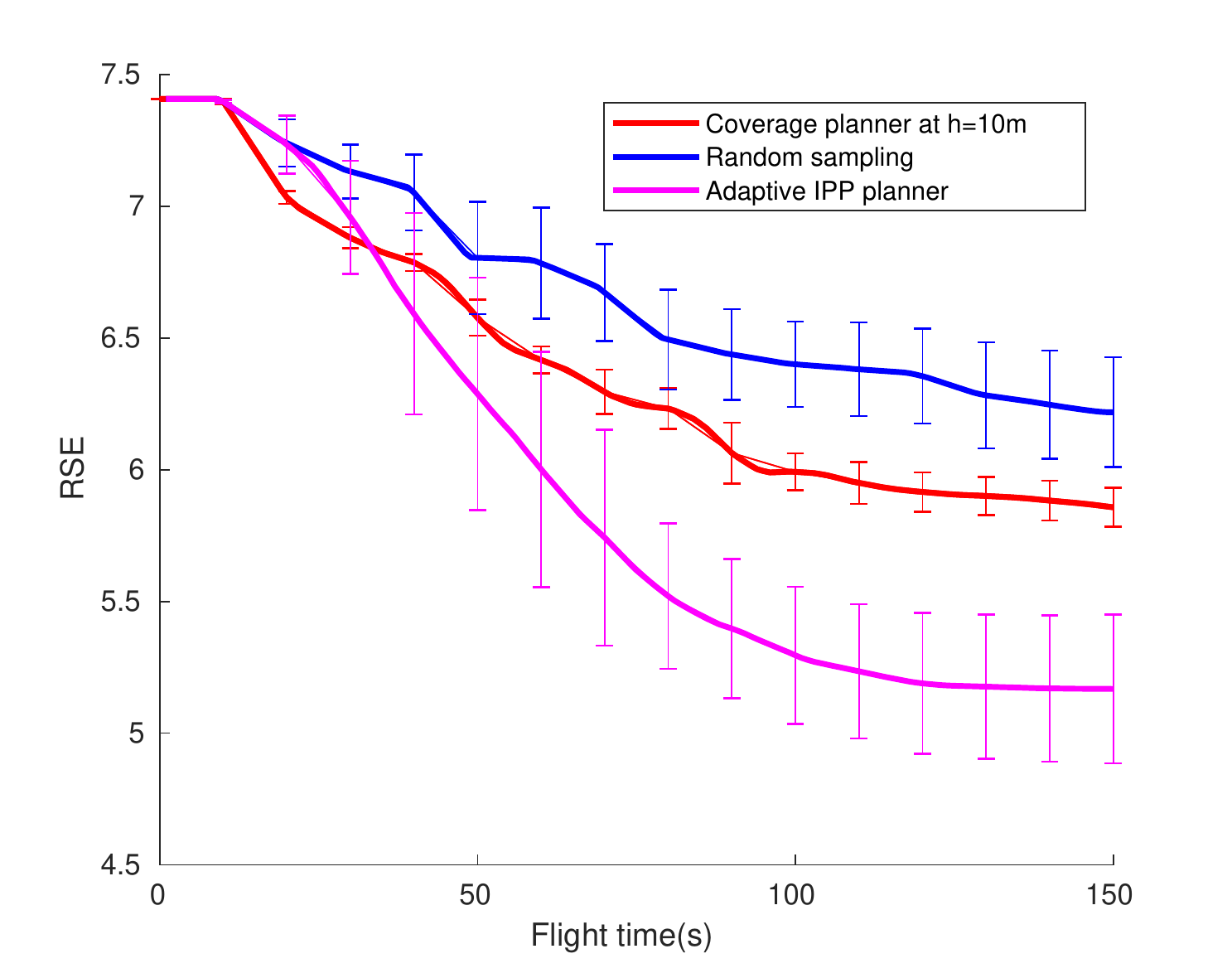}
\caption{Experimental results averaged over 25 flight trials shown with one standard deviation. Our planner (pink) outperforms coverage at altitude 10m (red) and random sampling (blue) by decreasing RSE at the fastest rate.}
\label{fig:bench_planners}
\end{SCfigure}

Figure \ref{fig:bench_planners} shows the evolution of RSE for each method during a mission.
Our OA-IPP method (pink) outperforms the state-of-the-art algorithms by reducing error at the fastest rate,
as it trades off between sensor performance and coverage at different altitudes,
and actively focuses on targets as they are found.
This strategy permits
target re-observation, which decreases false positive detections,
and variable resolution mapping, which refines accuracy over time.
In contrast,
the coverage benchmark (red) is limited in accuracy by the fixed altitude,
while the random planner performs worst
as it is often limited in coverage by the low altitude of sampled measurement sites.


\subsection{Adaptive replanning evaluation} \label{S:adaptive_planning_eval}

Next,
the effects of our adaptive replanning scheme are evaluated in terms of target search efficiency.
In the same simulation set-up as above,
we compare the performance of two variants of our algorithm
using the non-adaptive (Section~\ref{S:nonadaptive_planning}) and adaptive (Section~\ref{S:adaptive_planning}) planning objectives.
As before,
we run 25 trials and examine the evolution of RSE over time.
\begin{SCfigure}[][htpb]
\centering
\includegraphics[scale = 0.5]{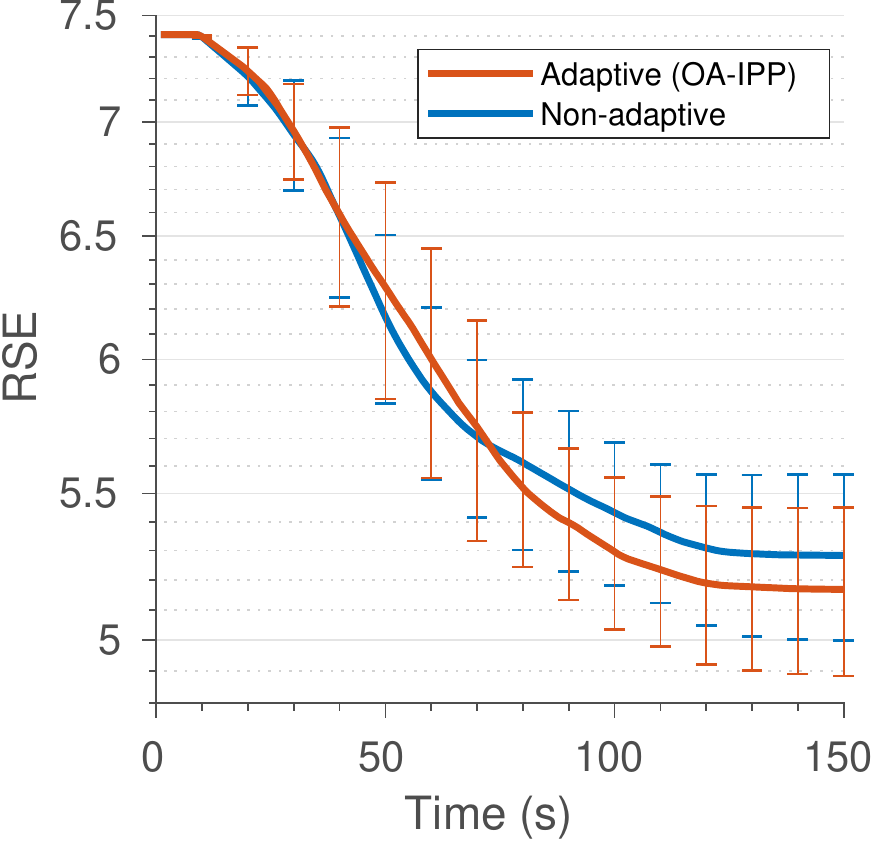}
\caption{Experimental results of 25 flight trials with one standard deviation as the error bar. Our adaptive planner (red) outperforms non-adaptive IPP planner (blue). Both prefers coverage strategy initially. However, towards the end, our planner generates more accurate map due to target re-observation strategy.}
\label{fig:bench_objective}
\end{SCfigure}

\begin{figure*}[!h]
    \centering
    \captionsetup{justification=justified}
     \begin{subfigure}[b]{0.23\textwidth}
		\includegraphics[width=\textwidth,trim={1.8cm 1.07cm 0 0},clip]{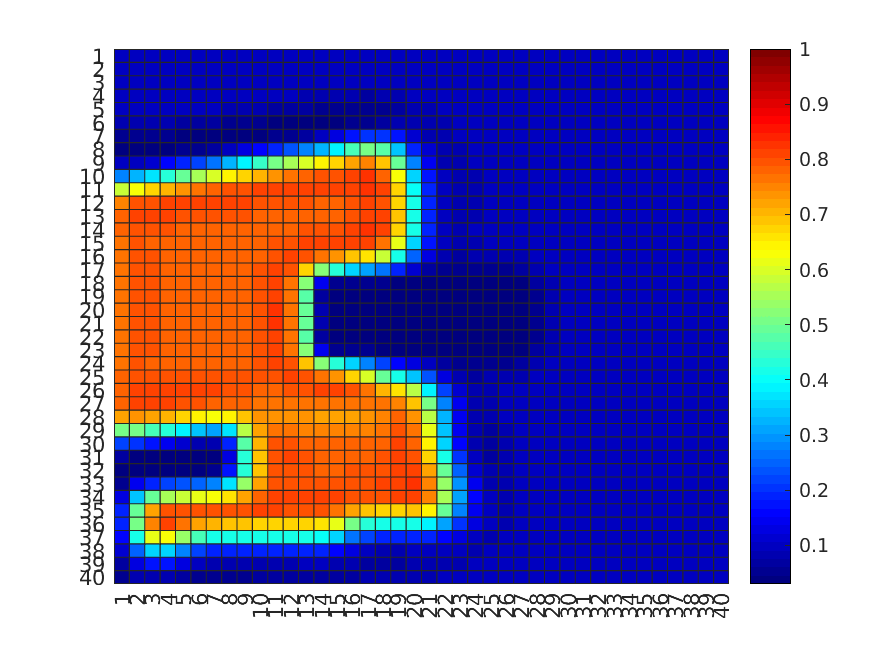}
    \end{subfigure} 
    \begin{subfigure}[b]{0.23\textwidth}
		\includegraphics[width=\textwidth,trim={1.8cm 1.07cm 0 0},clip]{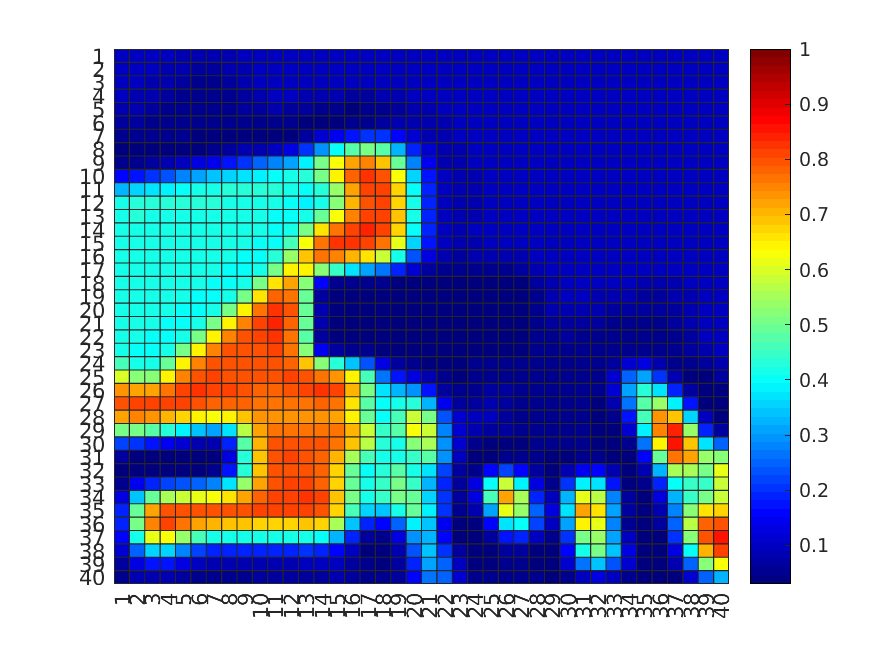}
    \end{subfigure} 
     \begin{subfigure}[b]{0.23\textwidth}
		\includegraphics[width=\textwidth,trim={1.8cm 1.07cm 0 0},clip]{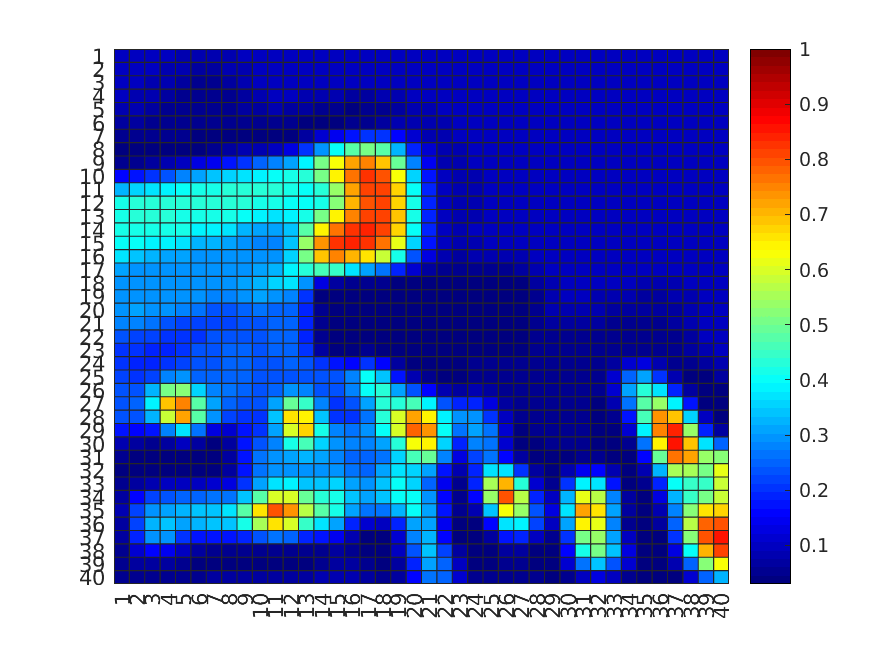}
    \end{subfigure} 
    \begin{subfigure}[b]{0.23\textwidth}
		\includegraphics[width=\textwidth,trim={1.8cm 1.07cm 0 0},clip]{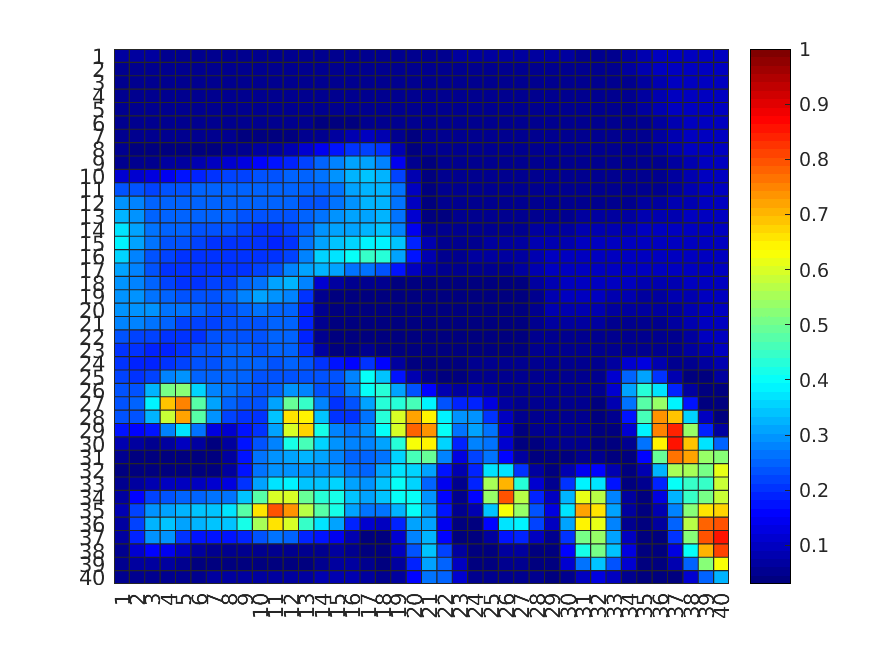}
    \end{subfigure}     
    \caption{GP map after 1st, 4rth, 7th and 11th measurement update. Our planner demonstrating its robustness against the false positive measurement (1st update) through target re-observation. All 7 humans are detected despite poor sensor performance.}
    \label{fig:robust_updates}
\end{figure*}

Figure \ref{fig:bench_objective} demonstrates that our OA-IPP algorithm outperforms non-adaptive IPP planner.
Initially, both the non-adaptive (blue) and adaptive (red) planner perform similarly
while the UAV performs field coverage (exploration).
At later stages of the mission ($< 75$\,s),
the adaptive variant achieves improved accuracy,
as the acquisition function permits
the UAV to focus on the half of the field where targets are likely to be found.
This behaviour leads to repeated target observations,
which renders the search robust to sensing uncertainty
and procures a higher-quality end map. 
This aspect illustrated in Figure \ref{fig:robust_updates},
which shows the GP mean after measurement updates for the environment in Figure \ref{fig:envt_men_one_side}.
Despite the false positive detection after the first update,
repeated measurements enable the recovery of a final map which resembles the ground truth in Figure \ref{fig:envt_men_one_side}.

\begin{figure}[htpb]
    \centering
    \begin{subfigure}[b]{0.23\textwidth}
		\includegraphics[width=\textwidth]{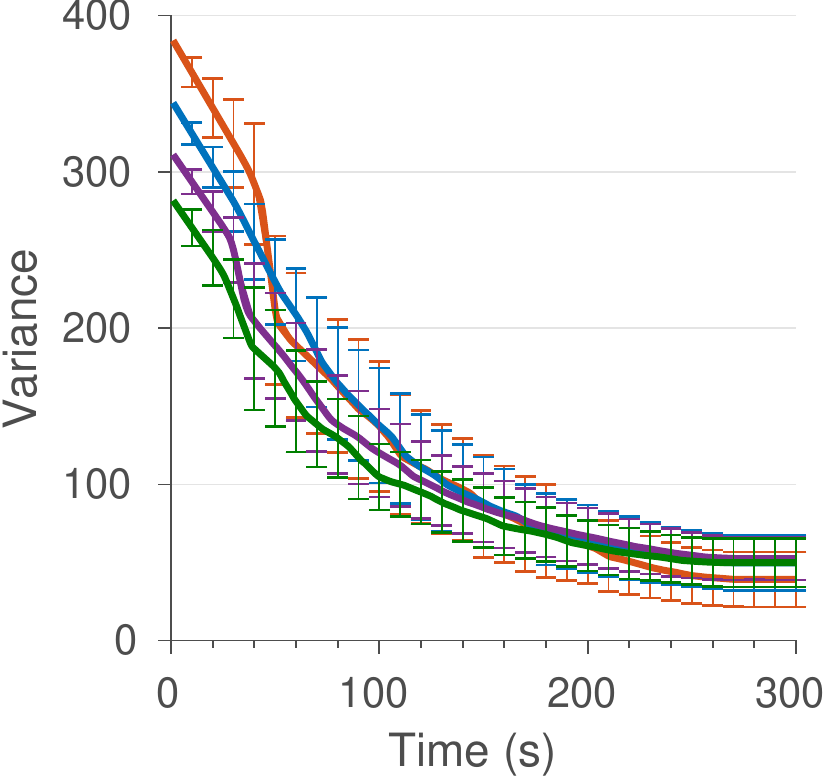}
		\caption{}\label{fig:rand_box_small_build}
    \end{subfigure} 
    \begin{subfigure}[b]{0.23\textwidth}
		\includegraphics[width=\textwidth]{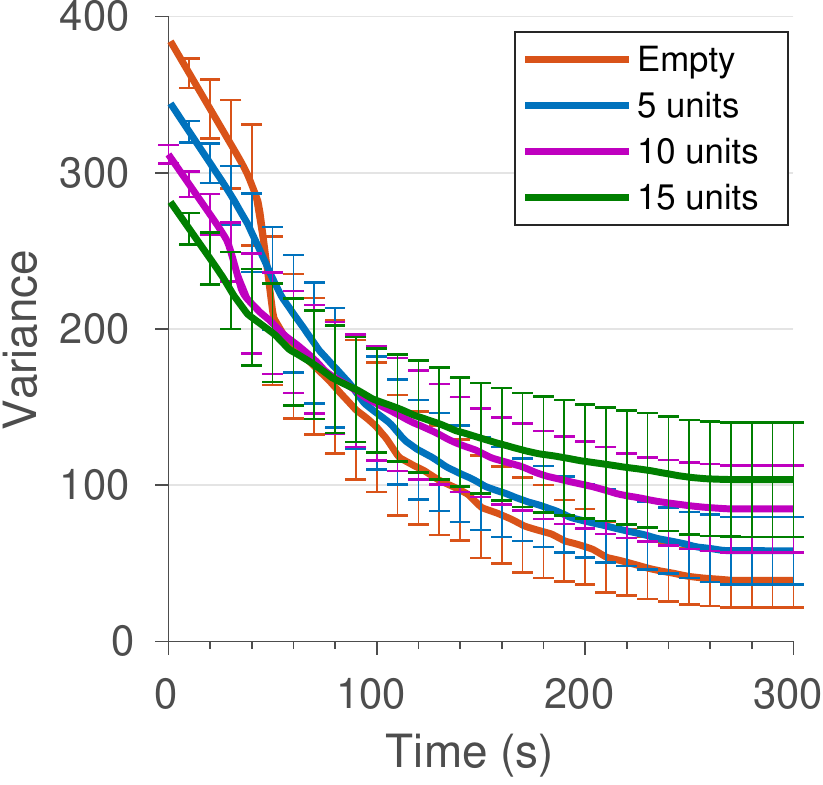}
		\caption{}\label{fig:rand_box_big_build}
    \end{subfigure}   
    \caption{Average results of our OA-IPP method in 100 flight trials for environments with low-rising (a) and high-rising (b) obstacles with varying densities.
    The error bars corresponds to one standard deviation.
    The scenario in (a) exhibits consistent performance due to improved FoV.
    However, in (b), high occlusion impedes exploration,
    leading to lower uncertainty reduction with increasing density.}\label{fig:rand_box_build}
\end{figure}

\subsection{Environment complexity} \label{S:envt_complexity}
Next,
the obstacle avoidance capabilities of our OA-IPP algorithm are assessed
by examining its performance in different environments.
We study two types of environment within a $30\times30$\,m area,
which contain (1) low-rising ($4\times4\times13$\,m) and (2) high-rising ($4\times4\times26$\,m) obstacles,
to portray SaR scenarios in different urban landscapes.
Case (2) is designed such that
the UAV must fly around, rather than over, obstacles
due to the saturation height $h_{sat}$ of the sensor.
To demonstrate the applicability of our approach,
we conduct $100$ trials in both scenarios,
considering variable obstacle densities
of $(5,10,15)$ units with randomly initialized positions.
Since exploratory capabilities are crucial for very occluded, complex environments,
we use uncertainty as the evaluation metric,
quantified by the covariance trace of the GP field model ($Tr(P)$).



\begin{figure}[htpb]
    \centering
     \begin{subfigure}[b]{0.17\textwidth}
		\includegraphics[width=\textwidth]{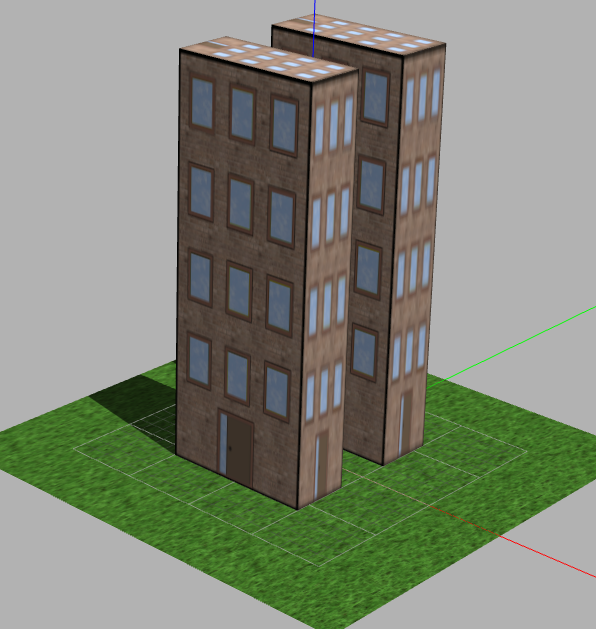}
		\caption{}
		\label{fig:envt_2_boxes}
    \end{subfigure} 
    \begin{subfigure}[b]{0.235\textwidth}
		\includegraphics[width=\textwidth]{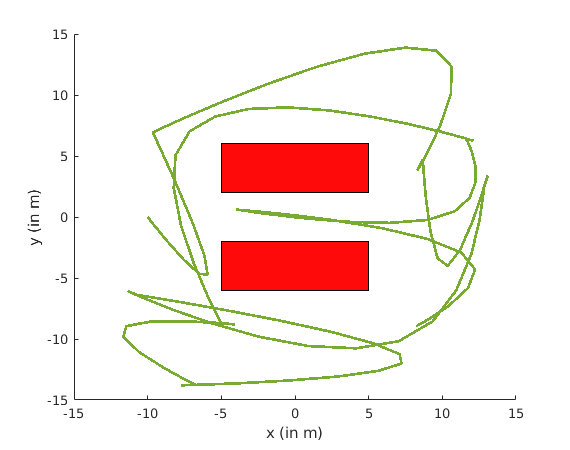}
		\caption{}
		\label{fig:path_planned_top_view}
    \end{subfigure} 
    \caption{Example of a collision-free trajectory planned using our approach (top view) (b) in a complex environment with 2 high-rising obstacles (a).}
    \label{fig:trajectory_2_buil}
\end{figure}

Our experimental results are depicted in Figure~\ref{fig:rand_box_build}.
Figure~\ref{fig:rand_box_small_build} confirms that
our planner achieves consistent uncertainty reduction with low-rising obstacles,
regardless of the environment density.
However, with an increased number of high-rising obstacles (Figure~\ref{fig:rand_box_big_build}),
variance reduction is limited due to increased occlusions of the camera FoV during mapping. 
As an example,
Figure \ref{fig:path_planned_top_view} shows the trajectory planned
in the narrow environment in Figure \ref{fig:envt_2_boxes} for a $150$\,s mission.
The flight path between the buildings validates that our approach
is capable of generating complex collision-free search trajectories
in challenging scenarios. Therefore, OA-IPP algorithm generalizes for a wide range of environment complexities and obstacle densities.

%

\subsection{Realistic simulation} \label{S:realistic_simulation}

Finally,
we validate our algorithm in a realistic RotorS-based simulation
of an urban SaR scenario.
Figures~\ref{fig:envt_real_iso} and \ref{fig:envt_real_top} show our experimental set-up
with 7 human targets for mapping.
Our aim is to demonstrate a single flight experiment in a more realistic urban environment,
and quantify the accuracy of the target map produced.
In this experiment,
we consider a $150$\,s mission
and apply the same mapping and planning parameters as in the previous sections.

\begin{figure}[htpb]
    \centering
    \begin{subfigure}[b]{0.2\textwidth}
		\includegraphics[width=\textwidth]{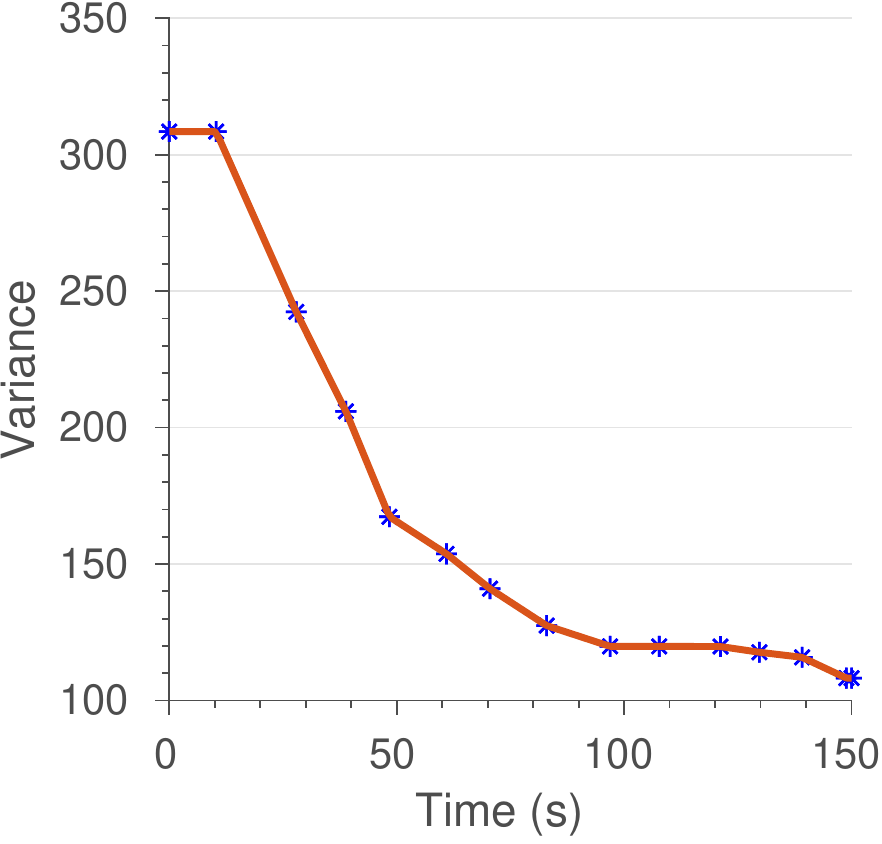}
    \end{subfigure} 
    \begin{subfigure}[b]{0.2\textwidth}
		\includegraphics[width=\textwidth]{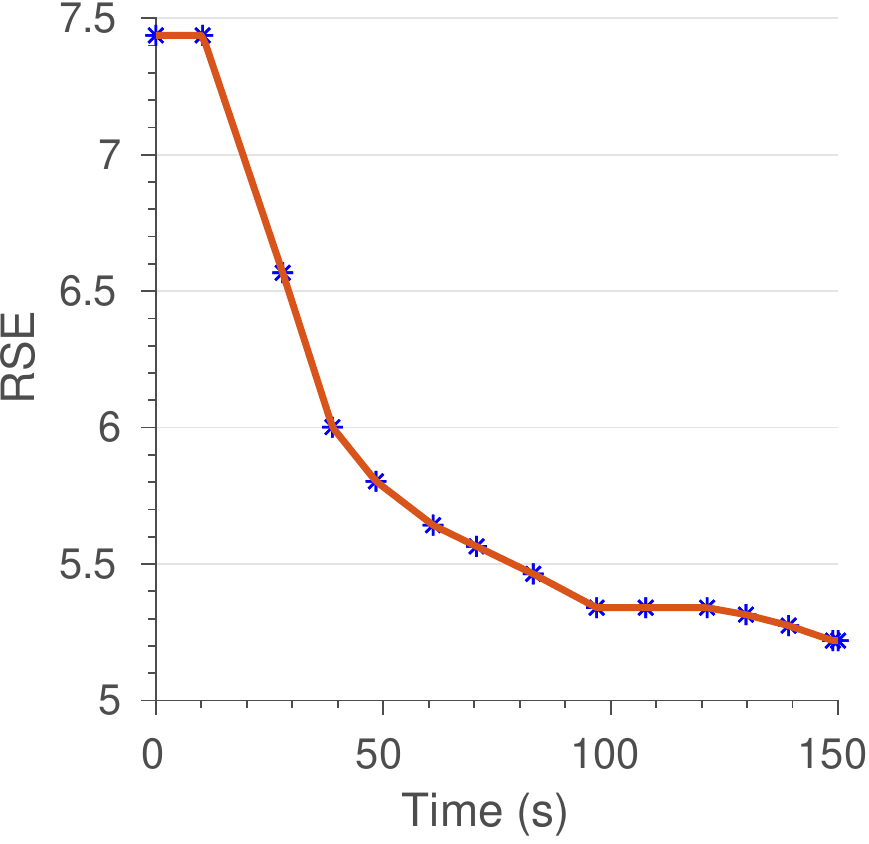}
    \end{subfigure}   
    \caption{Results for our realistic urban SaR mission.
    By planning adaptively,
    our OA-IPP methods achieves quick map uncertainty (left) and error (right) reductions
    for efficient target detection.}\label{fig:realistic_flight_perf}
\end{figure}

Figure \ref{fig:realistic_final_mean} depicts the final map result produced.
A visual comparison with the ground truth Figure \ref{fig:realistic_ground_truth}
confirms that our planner successfully finds the 7 human victims.
Quantitatively,
Figure \ref{fig:realistic_flight_perf} establishes that map uncertainty and error reduce
during the mission,
which implies that our method delivers an end result with increasing quality and confidence.
The map provides crucial information for practical applications, making the map quality and confidence highly important.

%
%
%

\section{CONCLUSIONS AND FUTURE WORK} \label{S:conclusion}
The paper introduced a novel obstacle-aware IPP algorithm that is applicable for target search problems using a UAV. The planner simultaneously trades off between coverage, obstacle avoidance, target re-observation, altitude dependent sensor performance, flight time and FoV to generate the optimal, finite-horizon 3D polynomial path in an obstacle filled environment. The proposed layered optimization approach facilitates a balanced exploration-exploitation strategy which makes it robust against false detections. Extensive simulations show that our planner outperforms non-adaptive IPP planner, coverage planner and random sampling planner in terms of search efficiency. The algorithm was generalized for a wide range of environment complexities and obstacle densities. It successfully found all the humans on the ground in a realistic SaR simulation, despite multiple false human detections by the sensor, demonstrating the robustness provided by the layered optimization approach.

The main drawback with the planner is the underlying assumption of a known and static environment. Moreover, a non-temporal field was assumed for target occupancy. Future research will investigate planning with dynamic obstacles and fields. Multi-UAV collaboration for large area search is also a promising direction for future work.

\section*{ACKNOWLEDGMENT}
This project has received funding from the European Union’s Horizon 2020 
research and innovation programme under grant agreement No 644227 and from the 
Swiss State Secretariat for Education, Research and Innovation (SERI) under 
contract number 15.0029.
It was also partly funded by the National Center of Competence in Research (NCCR) Robotics through the Swiss National Science Foundation.

\bibliographystyle{IEEEtranN}
\footnotesize
\bibliography{2019-anil-meera}

\begin{thebibliography}{25}
\providecommand{\natexlab}[1]{#1}
\providecommand{\url}[1]{#1}
\csname url@samestyle\endcsname
\providecommand{\newblock}{\relax}
\providecommand{\bibinfo}[2]{#2}
\providecommand{\BIBentrySTDinterwordspacing}{\spaceskip=0pt\relax}
\providecommand{\BIBentryALTinterwordstretchfactor}{4}
\providecommand{\BIBentryALTinterwordspacing}{\spaceskip=\fontdimen2\font plus
\BIBentryALTinterwordstretchfactor\fontdimen3\font minus
  \fontdimen4\font\relax}
\providecommand{\BIBforeignlanguage}[2]{{%
\expandafter\ifx\csname l@#1\endcsname\relax
\typeout{** WARNING: IEEEtranN.bst: No hyphenation pattern has been}%
\typeout{** loaded for the language `#1'. Using the pattern for}%
\typeout{** the default language instead.}%
\else
\language=\csname l@#1\endcsname
\fi
#2}}
\providecommand{\BIBdecl}{\relax}
\BIBdecl

\bibitem[Waharte and Trigoni(2010)]{waharte2010supporting}
S.~Waharte and N.~Trigoni, ``{Supporting search and rescue operations with
  UAVs},'' in \emph{International Conference on Emerging Security
  Technologies}.\hskip 1em plus 0.5em minus 0.4em\relax IEEE, 2010, pp.
  142--147.

\bibitem[Gupta et~al.(2017)Gupta, Bessonov, and Li]{gupta2017decision}
A.~Gupta, D.~Bessonov, and P.~Li, ``{A Decision-theoretic Approach to
  Detection-based Target Search with a UAV},'' in \emph{IEEE/RSJ International
  Conference on Intelligent Robots and Systems}.\hskip 1em plus 0.5em minus
  0.4em\relax IEEE, 2017, pp. 5304--5309.

\bibitem[Popovi{\'{c}} et~al.(2017)Popovi{\'{c}}, Vidal-Calleja, Hitz, Sa,
  Siegwart, and Nieto]{popovic2017multiresolution}
M.~Popovi{\'{c}}, T.~Vidal-Calleja, G.~Hitz, I.~Sa, R.~Siegwart, and J.~Nieto,
  ``{Multiresolution Mapping and Informative Path Planning for UAV-based
  Terrain Monitoring},'' in \emph{IEEE/RSJ International Conference on
  Intelligent Robots and Systems}.\hskip 1em plus 0.5em minus 0.4em\relax IEEE,
  2017.

\bibitem[Colomina and Molina(2014)]{Colomina2014}
I.~Colomina and P.~Molina, ``Unmanned aerial systems for photogrammetry and
  remote sensing: A review,'' \emph{ISPRS Journal of Photogrammetry and Remote
  Sensing}, vol.~92, pp. 79 -- 97, 2014.

\bibitem[Girard et~al.(2004)Girard, Howell, and Hedrick]{Girard2004}
A.~R. Girard, A.~S. Howell, and J.~K. Hedrick, ``Border patrol and surveillance
  missions using multiple unmanned air vehicles,'' in \emph{IEEE Conference on
  Decision and Control}, vol.~1, Dec 2004, pp. 620--625 Vol.1.

\bibitem[Linchant et~al.()Linchant, Lisein, Semeki, Lejeune, and
  Vermeulen]{Linchant2015}
J.~Linchant, J.~Lisein, J.~Semeki, P.~Lejeune, and C.~Vermeulen, ``{Are
  unmanned aircraft systems (UASs) the future of wildlife monitoring? A review
  of accomplishments and challenges},'' \emph{Mammal Review}, vol.~45, no.~4,
  pp. 239--252.

\bibitem[Geyer(2008)]{geyer2008active}
C.~Geyer, ``{Active target search from UAVs in urban environments},'' in
  \emph{IEEE International Conference on Robotics and Autmation}.\hskip 1em
  plus 0.5em minus 0.4em\relax IEEE, 2008, pp. 2366--2371.

\bibitem[Kurniawati et~al.(2008)Kurniawati, Hsu, and Lee]{Kurniawati2008}
H.~Kurniawati, D.~Hsu, and W.~S. Lee, ``{SARSOP : Efficient Point-Based POMDP
  Planning by Approximating Optimally Reachable Belief Spaces},'' in
  \emph{Robotics: Science and Systems}.\hskip 1em plus 0.5em minus 0.4em\relax
  Z{\"{u}}rich: MIT Press, 2008.

\bibitem[Popovi{\'c} et~al.(2017)Popovi{\'c}, Hitz, Nieto, Sa, Siegwart, and
  Galceran]{popovic2017online}
M.~Popovi{\'c}, G.~Hitz, J.~Nieto, I.~Sa, R.~Siegwart, and E.~Galceran,
  ``Online informative path planning for active classification using uavs,'' in
  \emph{IEEE International Conference on Robotics and Automation}.\hskip 1em
  plus 0.5em minus 0.4em\relax IEEE, 2017, pp. 5753--5758.

\bibitem[Binney et~al.(2013)Binney, Krause, and Sukhatme]{binney2013optimizing}
J.~Binney, A.~Krause, and G.~Sukhatme, ``Optimizing waypoints for monitoring
  spatiotemporal phenomena,'' \emph{The International Journal of Robotics
  Research}, vol.~32, no.~8, pp. 873--888, 2013.

\bibitem[Singh et~al.(2010)Singh, Ramos, Durrant-Whyte, and
  Kaiser]{singh2010modeling}
A.~Singh, F.~Ramos, H.~Durrant-Whyte, and W.~Kaiser, ``Modeling and decision
  making in spatio-temporal processes for environmental surveillance,'' in
  \emph{IEEE International Conference on Robotics and Automation}.\hskip 1em
  plus 0.5em minus 0.4em\relax IEEE, 2010, pp. 5490--5497.

\bibitem[Binney and Sukhatme(2012)]{binney2012branch}
J.~Binney and G.~Sukhatme, ``Branch and bound for informative path planning,''
  in \emph{IEEE International Conference on Robotics and Autmation}.\hskip 1em
  plus 0.5em minus 0.4em\relax IEEE, 2012, pp. 2147--2154.

\bibitem[Hollinger and Sukhatme(2014)]{hollinger2014sampling}
G.~A. Hollinger and G.~Sukhatme, ``Sampling-based robotic information gathering
  algorithms,'' \emph{The International Journal of Robotics Research}, vol.~33,
  no.~9, pp. 1271--1287, 2014.

\bibitem[Marchant and Ramos(2014)]{marchant2014bayesian}
R.~Marchant and F.~Ramos, ``Bayesian optimisation for informative continuous
  path planning,'' in \emph{IEEE International Conference on Robotics and
  Autmation}.\hskip 1em plus 0.5em minus 0.4em\relax IEEE, 2014, pp.
  6136--6143.

\bibitem[Brochu et~al.(2010)Brochu, Cora, and De~Freitas]{brochu2010tutorial}
E.~Brochu, V.~M. Cora, and N.~De~Freitas, ``A tutorial on bayesian optimization
  of expensive cost functions, with application to active user modeling and
  hierarchical reinforcement learning,'' 2010.

\bibitem[Dang et~al.(2018)Dang, Papachristos, and Alexis]{dang2018autonomous}
T.~Dang, C.~Papachristos, and K.~Alexis, ``Autonomous exploration and
  simultaneous object search using aerial robots,'' in \emph{IEEE Aerospace
  Conference}.\hskip 1em plus 0.5em minus 0.4em\relax IEEE, 2018.

\bibitem[Richter et~al.(2016)Richter, Bry, and Roy]{richter2016polynomial}
C.~Richter, A.~Bry, and N.~Roy, ``Polynomial trajectory planning for aggressive
  quadrotor flight in dense indoor environments,'' in \emph{The International
  Journal of Robotics Research}.\hskip 1em plus 0.5em minus 0.4em\relax
  Springer, 2016, pp. 649--666.

\bibitem[Oleynikova et~al.(2017)Oleynikova, Taylor, Fehr, Siegwart, and
  Nieto]{oleynikova2017voxblox}
H.~Oleynikova, Z.~Taylor, M.~Fehr, R.~Siegwart, and J.~Nieto, ``{Voxblox:
  Incremental 3D Euclidean Signed Distance Fields for On-Board MAV Planning},''
  in \emph{IEEE/RSJ International Conference on Intelligent Robots and
  Systems}, 2017.

\bibitem[Rasmussen and Williams(2005)]{Rasmussen:2005:GPM:1162254}
C.~E. Rasmussen and C.~K.~I. Williams, \emph{Gaussian Processes for Machine
  Learning}.\hskip 1em plus 0.5em minus 0.4em\relax MIT Press, 2005.

\bibitem[Redmon et~al.(2016)Redmon, Divvala, Girshick, and
  Farhadi]{redmon2016you}
J.~Redmon, S.~Divvala, R.~Girshick, and A.~Farhadi, ``You only look once:
  Unified, real-time object detection,'' in \emph{IEEE Conference on Computer
  Vision and Pattern Recognition}, 2016, pp. 779--788.

\bibitem[Hansen(2006)]{hansen2006cma}
N.~Hansen, ``{The CMA evolution strategy: a comparing review},'' in
  \emph{Towards a new evolutionary computation}.\hskip 1em plus 0.5em minus
  0.4em\relax Springer, 2006, pp. 75--102.

\bibitem[Hitz et~al.(2017)Hitz, Galceran, Garneau, Pomerleau, and
  Siegwart]{hitz2017adaptive}
G.~Hitz, E.~Galceran, M.-{\`E}. Garneau, F.~Pomerleau, and R.~Siegwart,
  ``Adaptive continuous-space informative path planning for online
  environmental monitoring,'' \emph{Journal of Field Robotics}, vol.~34, no.~8,
  pp. 1427--1449, 2017.

\bibitem[Cox and John(1992)]{cox1992statistical}
D.~Cox and S.~John, ``A statistical method for global optimization,'' in
  \emph{IEEE International Conference on Systems, Man, and Cybernetics}.\hskip
  1em plus 0.5em minus 0.4em\relax IEEE, 1992, pp. 1241--1246.

\bibitem[Furrer et~al.(2016)Furrer, Burri, Achtelik, and Siegwart]{Furrer2016}
F.~Furrer, M.~Burri, M.~Achtelik, and R.~Siegwart, ``{RotorS - A Modular Gazebo
  MAV Simulator Framework},'' in \emph{Robot Operating System (ROS): The
  Complete Reference (Volume 1)}, A.~Koubaa, Ed.\hskip 1em plus 0.5em minus
  0.4em\relax Cham: Springer International Publishing, 2016, pp. 595--625.

\bibitem[Galceran and Carreras(2013)]{galceran2013survey}
E.~Galceran and M.~Carreras, ``A survey on coverage path planning for
  robotics,'' \emph{Robotics and Autonomous Systems}, vol.~61, no.~12, pp.
  1258--1276, 2013.

\end{thebibliography}

\end{document}